\documentclass[letterpaper,11pt]{article}
\usepackage[utf8]{inputenc}
\usepackage[english]{babel}
\usepackage[T1]{fontenc}
\usepackage{listings}
\usepackage{graphicx}
\usepackage[margin=1in]{geometry}
\usepackage{amsmath}
\usepackage{amssymb}
\usepackage{tabularx}
\usepackage{biblatex}
\usepackage{rotating}
\usepackage{booktabs} 
\usepackage{rotating} 
\usepackage{multirow}
\usepackage{subcaption}
\usepackage{multicol}
\usepackage{hyperref}
\usepackage{longtable}
\usepackage{setspace}
\usepackage{tikz}
\usepackage{subcaption}

\hypersetup{
    colorlinks,
    linkcolor={blue!50!black},
    citecolor={blue!50!black},
    urlcolor={blue!50!black}
}

\usepackage{booktabs}

\makeatletter
\renewcommand{\maketitle}{%
  \null
  \vspace*{\fill}
  \begin{center}
    {\LARGE\@title\par}
    \vskip 1em
    {\large
      \lineskip .5em%
      \begin{tabular}[t]{c}%
        \@author
      \end{tabular}\par}%
    \vskip 1em
    {\large \@date}%
  \end{center}%
  \vspace*{\fill}
  \thispagestyle{empty}
  \clearpage
}
\makeatother

\title{\textbf{2024 NASA SUITS Report: LLM-Driven Immersive Augmented Reality User Interface for Robotics and Space Exploration}}
\author{Authors and Project Contributors: Kathy Zhuang\textsuperscript{1}, Zixun Huang\textsuperscript{1}, Yukun Song\textsuperscript{1},
\\
Rui Li\textsuperscript{1}, Yinuo Zhou\textsuperscript{1} 
        \\
        \\
        Project Contributers: 
        Yue Fan\textsuperscript{1}, Kai Mohl\textsuperscript{1}, Isaac Gonzalez\textsuperscript{1}, Yani Shi\textsuperscript{1},
        \\
        Xiaowen Yuan\textsuperscript{1}, Rojan Kashani\textsuperscript{1}, Yunting Zhao\textsuperscript{1},  Aarav Desai\textsuperscript{1}, Ivan Apodaca\textsuperscript{1},
        \\
        Advika Govindarajan\textsuperscript{2}, Ishaan Patel\textsuperscript{2}, Colin Strout\textsuperscript{2}, Caleb Strout\textsuperscript{2}, Emily Guo\textsuperscript{2},
        \\
        Aeriel Amparo\textsuperscript{1}, Patrick Lee\textsuperscript{1}, Yaamini Jois\textsuperscript{1}, Patrick Lasiter\textsuperscript{2}
        \\
        \\
        Supervisors: Allen Y. Yang\textsuperscript{1}, Sheri L. Dragoo\textsuperscript{2}
        }
        
\date{June 2024}

\begin{document}
\textsuperscript{1}University of California, Berkeley \\
\textsuperscript{2}Baylor University \\
\texttt{\{kathy\_zhuang, zixun, yukun\_song, allenyang\}@berkeley.edu}

\maketitle

\pagenumbering{gobble}
\doublespacing

\clearpage

\tableofcontents

\clearpage

\pagenumbering{arabic}
\setcounter{page}{1}

\section{Project Overview}
As modern computers evolve to have better performance and larger computation power, new modes of interaction emerge along with novel designs of devices. Augmented Reality (AR), which allows users to interact with common objects via virtual interfaces, establishes a new space of designs and development and hence poses a significant challenge to machines' spatial perception of the complex real-world environment. Pose estimation of 3D objects, as one of the most commonplace and important tasks in 3D perception, is still a challenging task for modern algorithms, especially in complex real environments with alternating scene properties and irregular object geometries.

Our project is at the forefront of addressing critical challenges in human-robot interaction within dynamic mobile AR environments. We focus on exploring the potential ways of interacting with robots in space, especially in a non-intrusive manner. This interaction is made possible by integrating three key components: a non-intrusive head-mounted device serving as a user interface, voice control to enable astronauts to manipulate the interface and interact with robots using verbal commands, and a robot tracking algorithm that accurately localizes the robot's position in 3D space. Enabled by these technologies, we proposed URSA, an LLM-driven immersive AR user interface for robotics and space exploration, and participated in the 2023 NASA Spacesuit User Interface Technologies for Students (NASA SUITS) \cite{NASA_2024}. This project aims to develop solutions for future spaceflight needs, particularly for the Artemis missions, which seek to establish a sustained human presence on the Moon and Mars. The NASA SUITS project focuses on developing user interfaces for spacesuits and mission control consoles, which are critical for the effective operation of rovers and other equipment on the Martian surface.

\begin{figure*}
    \centering
    \includegraphics[width=1\linewidth]{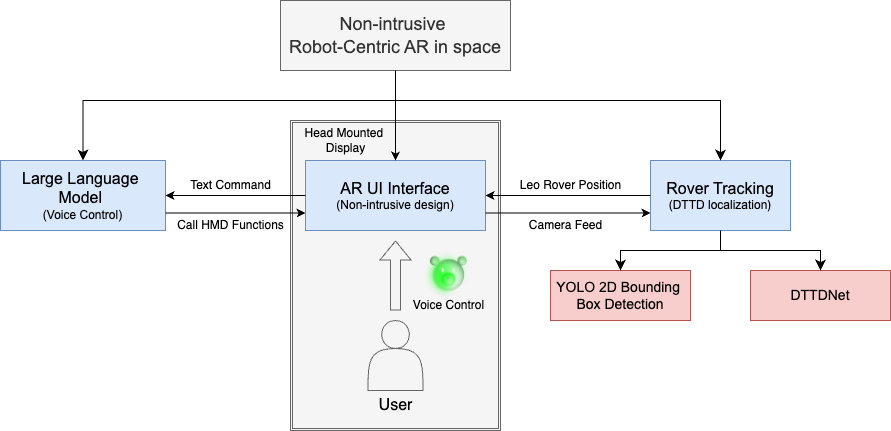}
    \caption{Structure of Augmented Reality Control Flow}
    \label{fig:flow}
\end{figure*}

Industries that rely on robotics to perform complex tasks are the main stakeholders. Researchers in the fields of computer vision, robotics, and augmented/virtual reality have also invested heavily in similar challenge for object tracking in AR dynamic space. In a human-robot interaction setting, especially when dangerous or complex tasks are involved, there has been needs for intuitive interactions and precision. 

To foster intuitive interaction, we have explored multiple aspects of the user interfaces within AR devices. Display technologies, serving as the visual output for these devices, have seen extensive development, evidenced by pioneering products such as the Microsoft HoloLens \cite{Microsoft_Hololens}, Google Glass \cite{Gvora_2023}, and HTC Vive XR Elite \cite{HTC_Vive}. These technologies offer a spectrum of possibilities for presenting information to users. On the input side, modalities like gaze tracking, hand tracking, voice control, and physical controllers have been rigorously investigated for their efficacy in interfacing with AR devices. This paper focuses on non-intrusive AR interfaces by integrating the discreet visual output capabilities of the HoloLens \cite{Microsoft_Hololens} with the sophisticated interaction offered by large language models (LLMs) for voice control. 

To achieve precision, we used digital twin localization technology. For example, in the early versions of the Digital Twin Tracking Dataset (DTTD) \cite{feng_digital_2023}, and DTTD2 \cite{huang_robust_2023}, have set the groundwork for data collection under varied conditions and introduced algorithms that improve accuracy amidst real-world data noise.  Similarly, by utilizing digital twin technology, we control robots via AR headset to perform accurate and dangerous tasks. Through such a control, the robot is capable of moving around to carry out work, while its position and poses being captured by external sensors and reflected on the AR headset simultaneously, enabling real-time remote control. This is particularly useful when robots need to be tracked with high accuracy without ground truth sensors. Our ongoing work emphasizes addressing the challenges of noise variance and real-time tracking accuracy with specialized camera systems like the ZED2, positioning our research at the forefront of digital twin applications in robotics. This focus provides novel insights and solutions in a rapidly evolving field.  The development of robust datasets and algorithms for digital twin technology not only advances technical capabilities but also revolutionizes industrial processes, with a focus on industries that utilize complex robotics. 

Our specific solutions for realizing these are as follows:

\hspace{0.5in} 1) We implemented an intuitive AR user interface using non-intrusive display and LLM-based voice control.

\hspace{0.5in} 2) We created a comprehensive robot-specific dataset using the ZED2 camera, specifically designed for non-rigid bodies. 

\hspace{0.5in} 3) We designed an local Mission Control Console (LMCC) user interface to provide intuitive and real-time monitor of the space exploration task. 

\hspace{0.5in} 4) We deploy the dataset to the transformer-based 6DoF pose estimator (DTTDNet) with depth-robust designs on modality fusion and training strategies to enable real-time, high-level tracking accuracy. 

\hspace{0.5in} 5) We apply our datasets, model tracking, and associated technologies within the industry to fulfill the requirements of astronauts' missions, and developed a comprehensive AR user interface, as demonstrated in Figure \ref{fig:flow}.

\clearpage
\section{Non-Intrusive AR User Interface for Head-Mounted Displays}

In the demanding and dynamic environment of space operations, the user interface for augmented reality (AR) systems, especially for astronauts, must prioritize minimal cognitive load and high efficiency. Astronauts require a system that maintains their awareness and focus on their surroundings, rather than diverts attention to manual controls or complex navigation menus. Our solution harnesses voice control to manage the AR interface non-intrusively, enhancing the interaction without overwhelming the user.

\subsection{Non-Intrusive Design}
\begin{figure*}
    \centering
    \includegraphics[width=0.8\linewidth]{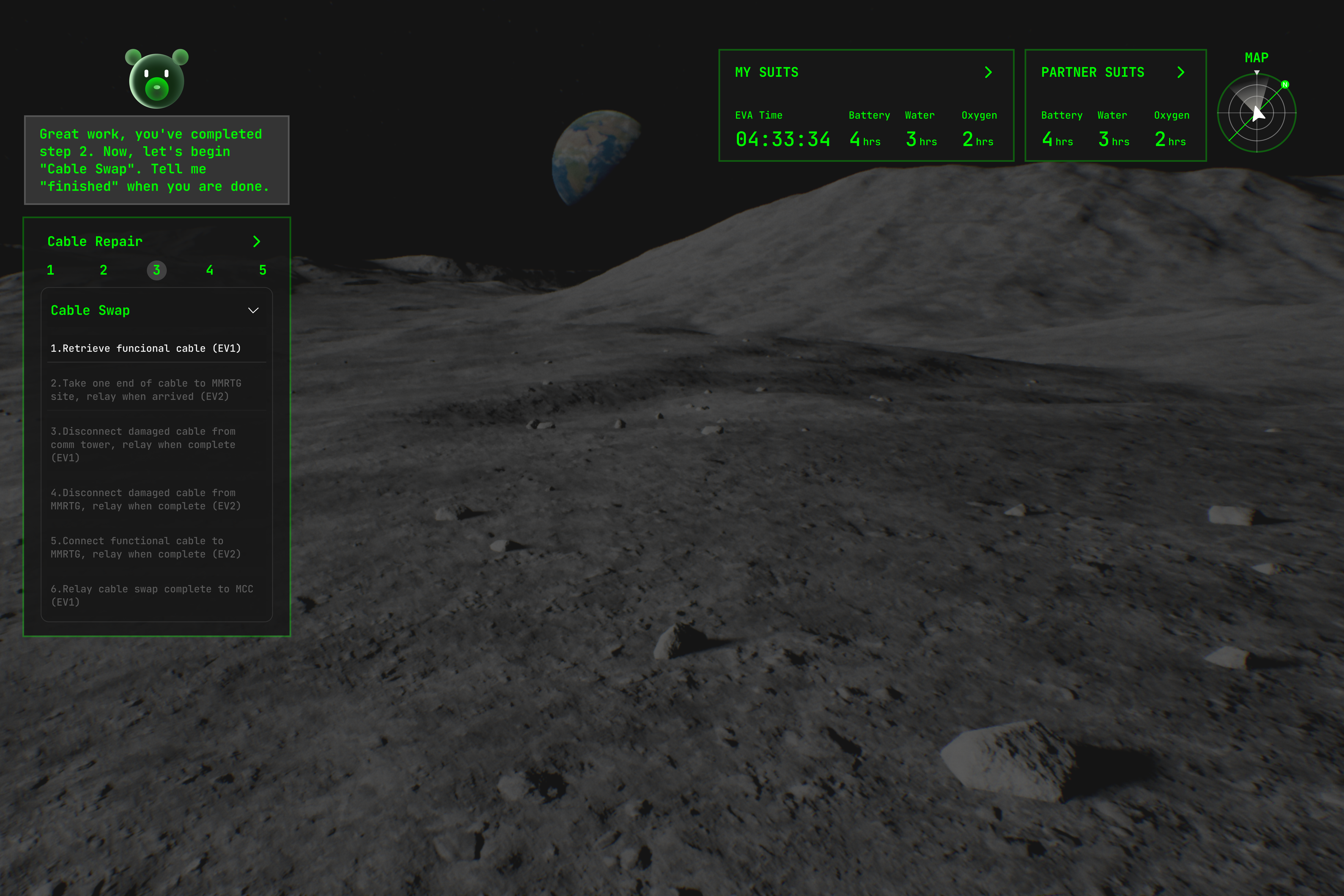}
    \caption{Non-intrusive design for HMD (Showing the Suits Panel, EVA Task Panel and Virtual Voice Assistant -- Ursa}
    \label{fig:nonintrusive}
\end{figure*}

In the context of space exploration, where environmental awareness and visual clarity are paramount, our AR interface is designed to be non-intrusive, prioritizing the astronaut's unobstructed view of their surroundings (Figure \ref{fig:nonintrusive}). Traditional AR interfaces often employ methods like gestures, touch pads, or gaze tracking. While these are effective on Earth, in the unique conditions of space, they can complicate interactions by cluttering the visual field or requiring manual inputs that are impractical in a bulky spacesuit.

\subsection{Integration with Head-Mounted Display (HMD)}
For our augmented reality user interface, we selected the Hololens 2  \cite{Microsoft_Hololens} as the preferred head-mounted display (HMD) due to its specific features that cater to the safety and operational requirements of astronauts. The Hololens 2 \cite{Microsoft_Hololens} stands out among various HMD options for several critical reasons, particularly its non-intrusive design and fail-safe visibility features, which are essential in the challenging environment of space. One of the paramount concerns in designing AR interfaces for space is ensuring that astronauts maintain an unobstructed view of their surroundings. The Hololens 2 \cite{Microsoft_Hololens} is designed with a transparent visor, allowing users to see their environment directly, with digital overlays enhancing but not obscuring their real-world view. This transparency is crucial in space, where situational awareness is vital for safety and efficiency. Importantly, even if the Hololens 2 \cite{Microsoft_Hololens} experiences a failure or powers down, the astronaut's view remains unblocked, ensuring continuous visibility of their surroundings, which is a critical safety feature not provided by all HMDs.

Our User Interface presents necessary data without overlaying directly over the astronaut’s main field of view. Information is displayed in peripheral areas or in a minimally invasive manner, ensuring that the most critical sightlines are always clear. The system is designed to respond to a set of predefined voice commands, which can control the display's configuration, toggle through data screens, or activate tasks without manual input.

Our project also encompasses comprehensive product management processes. This includes the development of prototypes ranging from low-fidelity to high-fidelity using Figma, UI development in Unity, and establishing backend server connections. Furthermore, we have employed techniques such as user experience mapping and lean startup market testing to gain a deep understanding of astronaut needs for Mars missions.

\section{Voice-Controlled AR User Interfaces}

\subsection{Traditional AR Controls And Their Limitations }
In the field of AR traditional control methods such as hand gestures, eye gaze, and handheld controllers are used to enhance user interaction. These methods are pivotal for improving the immersiveness and intuitiveness of AR environments -- Hand gestures are utilized for direct manipulation in virtual environments, offering a natural way to interact without the need for physical contact  \cite{9283348}; Eye gaze detection provides a means for selecting and activating items simply by looking at them, making interactions quicker and reducing the physical effort required \cite{DBLP:journals/corr/abs-2108-05479}; Handheld controllers continue to be relevant due to their precision and the tactile feedback they provide, which are crucial for complex interactions that require precision. Haptic controllers also offer physical shapes to the user to experience a sense of touch, including sounds, pressure, and vibrations \cite{physical_controller}. 

These traditional controls are foundational in AR technologies, supporting a range of applications from gaming and virtual training to medical and educational uses. However, in the realm of space exploration, these methods face significant practical limitations. For example, astronauts often need their hands free for critical tasks such as sample scanning and rover control, making hand-based interaction methods impractical. Additionally, the need for astronauts to continuously observe their surroundings means that using gaze tracking could undesirably restrict their ability to move their gaze freely. Furthermore, the bulky design of space suits limits astronauts' mobility, complicating the use of handheld controllers and reducing the precision of gesture and eye-tracking systems. Most critically, the cognitive demands of operating in space necessitate interaction methods that minimize cognitive load, favouring simpler and more robust control mechanisms that enhance intuitiveness.

\subsection{Voice-Controlled Interfaces Leveraging Large Language Model}

In response to the limitations of traditional AR control methods in space exploration, voice-controlled interfaces leveraging large language models (LLMs) present a promising alternative. These interfaces utilize advanced natural language processing technologies to interpret and execute commands spoken by astronauts and manipulate the user interfaces following the command. Such a method significantly reduces the physical interactions required. At the same time,  voice commands can be highly intuitive and adaptable to various operational contexts, it ensures minimum cognitive load and allows astronauts to maintain their focus on critical tasks. By integrating voice recognition systems, astronauts can perform complex operations efficiently, even while wearing bulky space suits or when their hands are occupied with other tasks. 

Recent advancements in function-calling capabilities have significantly enhanced the integration of voice-controlled interfaces in AR technologies through the use of large language models (LLMs). Gorilla, a model fine-tuned on the LLaMA architecture, stands out in this regard by expertly crafting API calls \cite{patil2023gorilla}. This model not only addresses the shortcomings of its previous works like GPT-4 but also excels in generating precise API calls while reducing common issues such as hallucinations \cite{patil2023gorilla}. Equipped with an integrated document retriever, Gorilla adeptly adapts to changes in documentation at test time, ensuring that API calls remain accurate and current \cite{patil2023gorilla}. Such advancements are crucial for dynamic and precise tool usage in AR environments, especially in contexts like space exploration where traditional physical controls are impractical.

\begin{figure}
    \centering
    \includegraphics[width=0.75\linewidth]{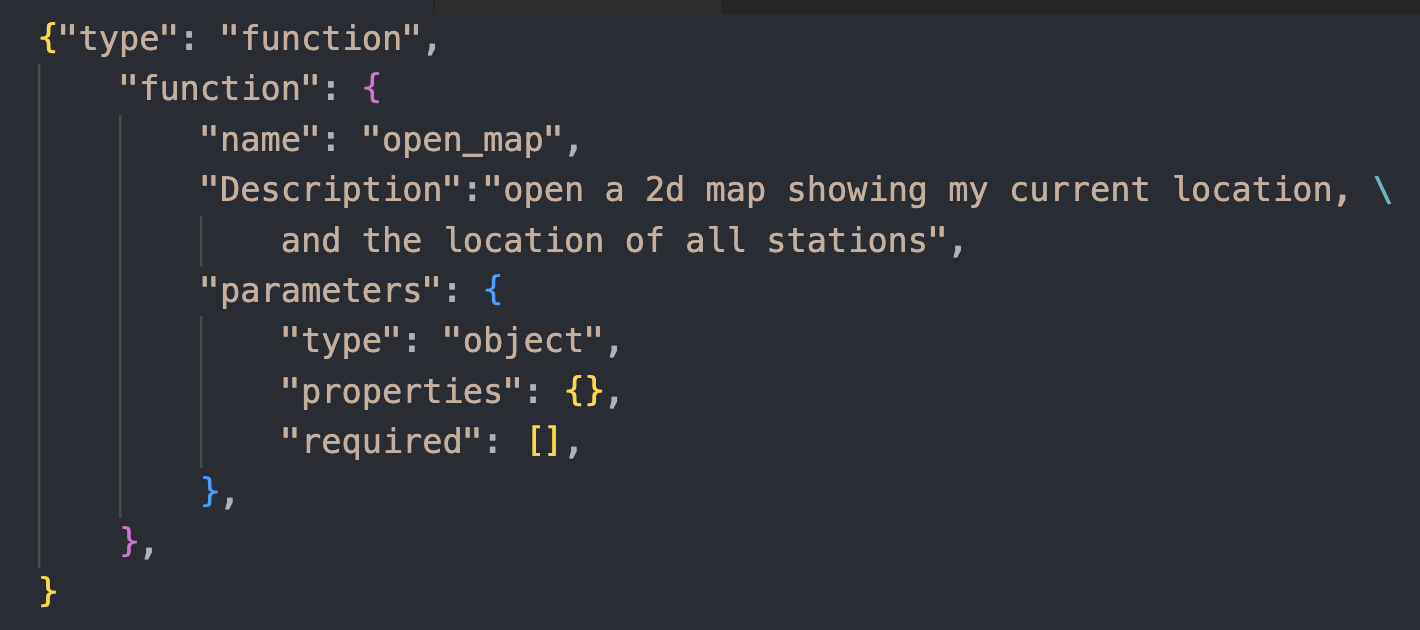}
    \caption{Function Description Example}
    \label{fig:function_description}
\end{figure}

Gorilla is seamlessly integrated into the AR user interface, which streamlines the process from issuing a voice command to performing corresponding command on the user interfaces. For example, as illustrated in Figure \ref{fig:flow}, an astronaut might issue a command like \textit{"Ursa, open the map for me"} (Ursa is the cue word for voice command) to their Hololens,  which is equipped with voice recognition capability. The Hololens converts the spoken command to text using its voice-to-text function and forwards text \textit{"open the map for me"} to Gorilla. Gorilla then processes the text command and, using its function-calling ability, executes the open\_map() function. Gorilla picks the function referencing to the function descriptions stored in OpenAI function calling format as shown in \ref{fig:function_description}. The function call then triggers the Hololens to display a 2D map to the user. 

\begin{figure}
    \centering
    \includegraphics[width=1\linewidth]{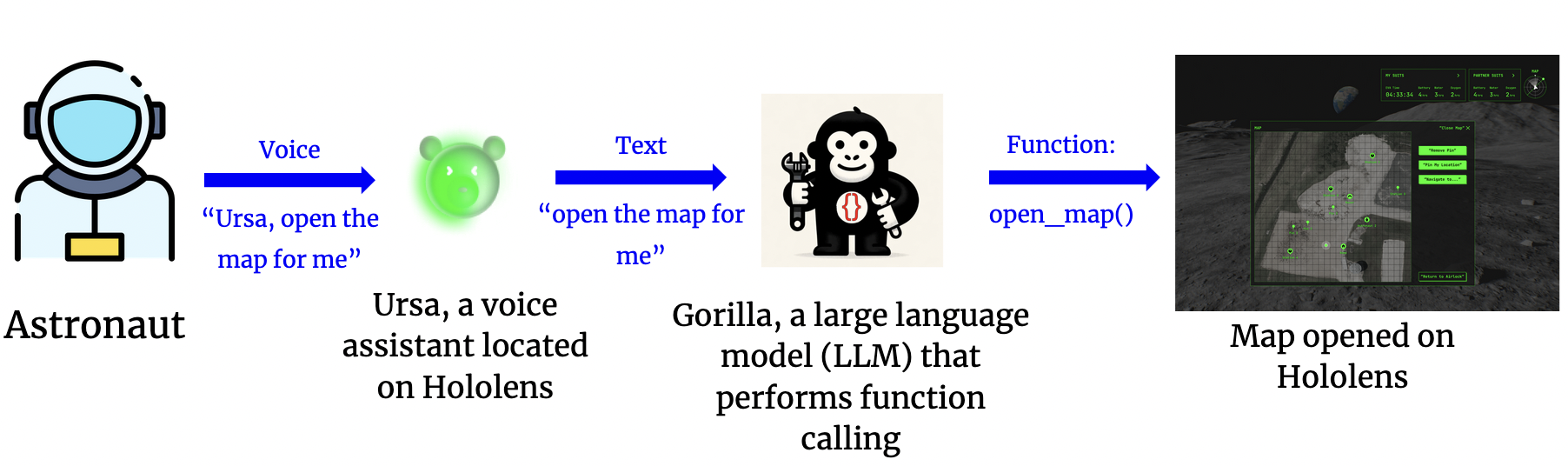}
    \caption{Flow of Voice-Controlled User Interface in AR}
    \label{fig:flow}
\end{figure}

This intuitive user interface, which effectively bridges the gap between verbal commands and complex digital interactions, exemplifies the potential of LLMs like Gorilla in enhancing user experiences in technology-augmented environments.

\section{Local Mission Control Console User Interface}
\begin{figure}
    \centering
    \includegraphics[width=1\linewidth]{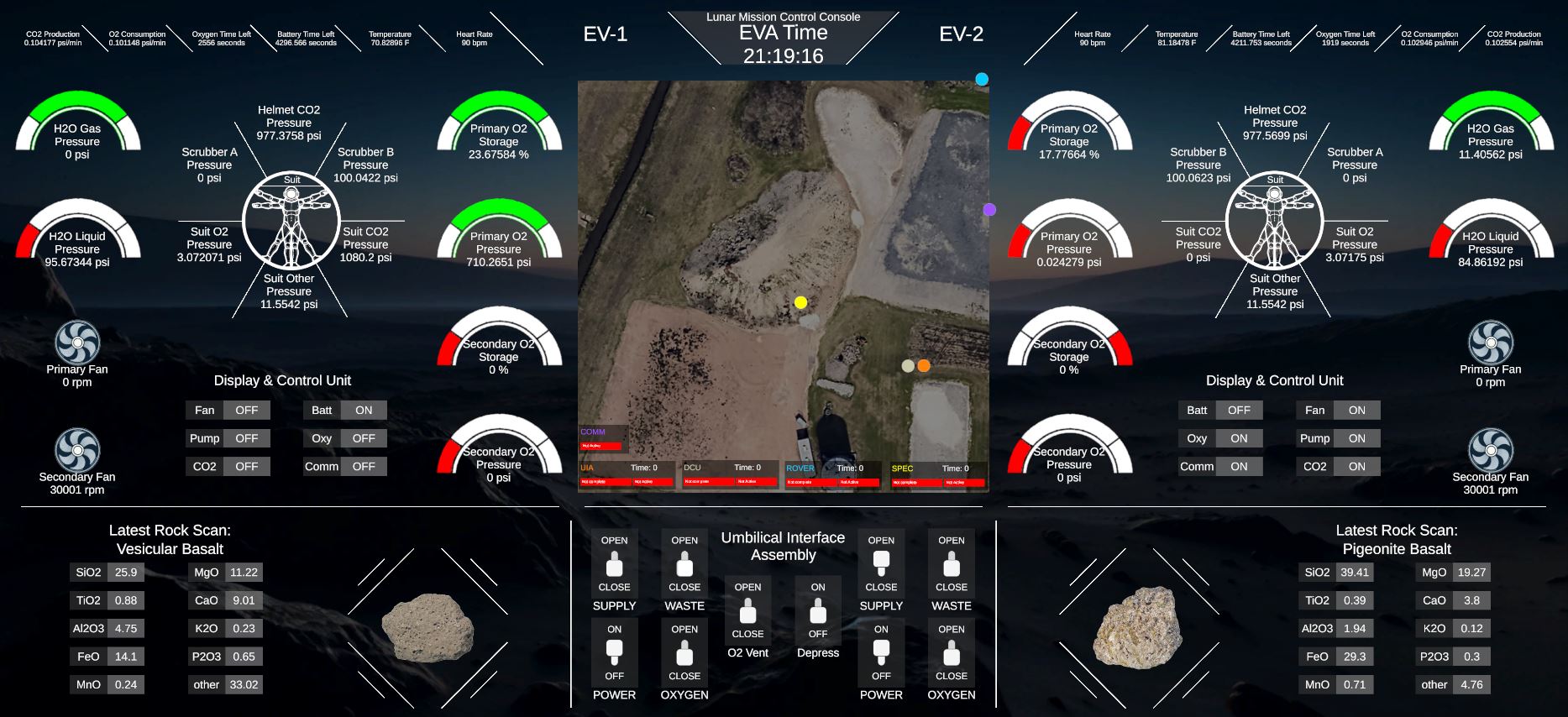}
    \caption{LMCC UI}
    \label{fig:LMCC}
\end{figure}

The LMCC interface effectively integrates  critical features to support space missions. As \ref{fig:LMCC} shows, it prominently displays essential biomedical data, such as heart rate and oxygen levels, and detailed spacesuit system information, including helmet CO2 and suit oxygen pressures. The interface also includes a live 2D map for tracking crewmembers and rover locations, with functionality for dropping and syncing customizable pins across different platforms. While it displays scientific data like rock scan compositions and supports task management with mission timers, the UI can be enhanced by adding rover system state data, remote piloting capabilities, and a caution and warning system for monitoring off-nominal telemetry. Additionally, the integration with Hololens 2 head-mounted displays allows astronauts to view critical data without obstructing their field of view, ensuring continuous situational awareness. To further improve the interface, features such as real-time task procedure displays and seamless data storage for scientific information could be integrated.

\section{Rover Tracking for AR User Interfaces}
Another vital functionality of the AR User Interface is real-time rover location tracking with visualization on the Head-Mounted Display (HMD). This task encounters two primary challenges: firstly, there is a lack of 3D object tracking datasets that specifically feature robots; secondly, the development and availability of object tracking algorithms tailored for rovers are limited. Addressing these issues is essential for enhancing the functionality and reliability of AR systems used in rover navigation and monitoring.

The core idea for precising tracking is to develop the a dataset that can effectively handle the variance in noise across different data types. The dataset will be curated through camera calibration, pose estimation, and annotation using OpenCV technology. We analyze our progress and project outcomes using clear data metrics, such as field-standard metrics of average point distance (ADD) and average closest point distance (ADD-S) to evaluate the performance of the model. In parallel, digital-twin technology serves as a virtual representation of an object or system, continually refreshed with real-time data while leveraging simulation and machine learning capabilities. 

\begin{figure}
    \centering
    \includegraphics[width=0.8\linewidth]{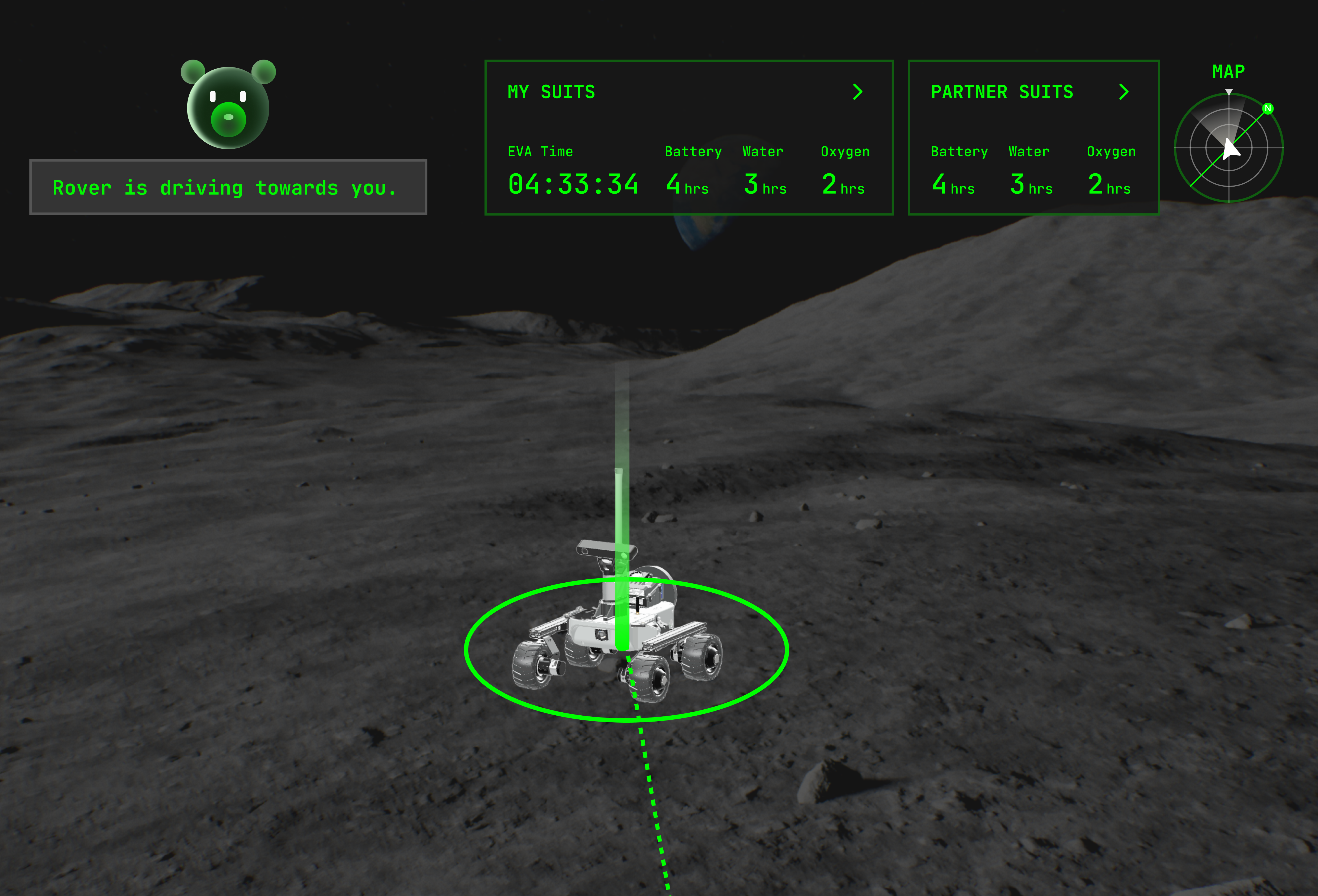}
    \caption{NASA SUITS HMD UI with Rover Tracking Overlay.}
    \label{fig:nasa_suits}
\end{figure}

When applying the rover tracking to NASA SUITS, we overlay a 2D circle highlight to indicate the precise position of the Leo Rover in the AR user interface (Figure \ref{fig:nasa_suits}).

\subsection{Rover Tracking Dataset}

\subsubsection{The novelty of DTTD3}

The development of 3D object tracking technologies has greatly benefited from synthetic and real-world datasets, which provide essential data for training object detection algorithms. Synthetic datasets like ShapeNet6D, the synthetic dataset from Hinterstoisser, and the FAT dataset, offer photo-realistic images that are easily accessible and require less human effort for annotation, yet they may suffer from domain shift issues that affect model robustness in real scenarios \cite{he_fs6d_2022, hinterstoisser_model_2013, tremblay_falling_2018}.

Real-world datasets, gathered through various sensors including stereo cameras and structured light devices, provide valuable data directly from real-life conditions. These datasets, like YCB-Video and TOD, are collected using complex setups to ensure the capture of detailed and accurate object poses but require extensive manual annotation and are subject to inaccuracies from the capturing devices \cite{xiang_posecnn_2018, liu_keypose_2020, kaskman_homebreweddb_2019}.

Identifying the need for pose estimation for non-rigid-body objects, we collected the DTTD3 specifically for robots with movable parts under several conditions. Utilizing a motion capture system as the anchor of ground truth coordinates, we collected a total of 18 scenes of approximately 5,000 labelled images. The scenes all contain Leo Rover, each with a different pose and orientation, a distance of 0.5m, 1m, or 2m from ZED camera, and the rover being placed at 0.5m or 1m. To enhance the dataset for training and validation purposes, we further enhanced the dataset with 30,000 synthetic images using the 3D ground truth model collected either from the manufacturer or via Polycam. Below, we illustrate the data collection process in detail.

\subsubsection{DTTD3 Data Collection Pipeline}
\begin{figure}
    \centering
    \includegraphics[width=0.8\linewidth]{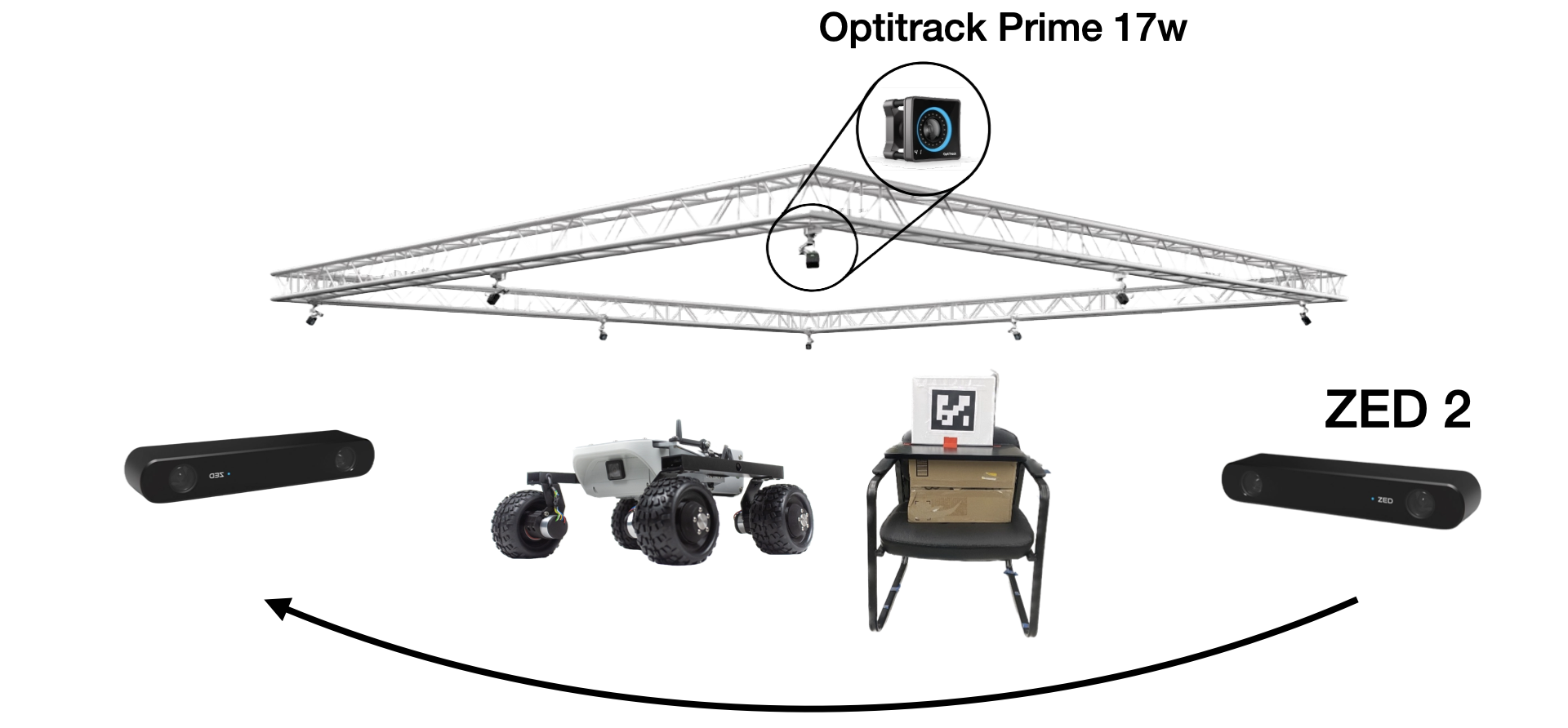}
    \caption{The conceptual setup for collecting the Digital-Twin Tracking-Dataset. A ZED-2 camera is utilized to capture RGB-D data from robots (ex. the Leo Rover) from multiple angles. The ARUCO marker is utilized as an initial calibration of the ZED-2 camera extrinsic matrix before scene collections and is occluded in every scene.}
    \label{fig:mocap}
\end{figure}

The DTTD3 data collection pipeline utilizes 10 OptiTrack Prime 17W infrared cameras to track specialized infrared reflective markers to establish the 6D ground truth pose of the camera frame, alongside a Stereolabs ZED2 RGB-D camera for capturing high-resolution images and depth data. This setup, including multiple robots with varied features, is calibrated using an ARUCO marker to align the OptiTrack and ZED2 camera frames. (Figure \ref{fig:mocap})

Intrinsic camera calibration is conducted using OpenCV, and the transformation between the OptiTrack and ZED2 spaces is determined through measurements on the ARUCO marker. Depth images, paired with calibrated intrinsics, allow for the creation of a dense 3D point cloud which is aligned with the global OptiTrack coordinates.
\begin{figure}
    \centering
    \includegraphics[width=\linewidth]{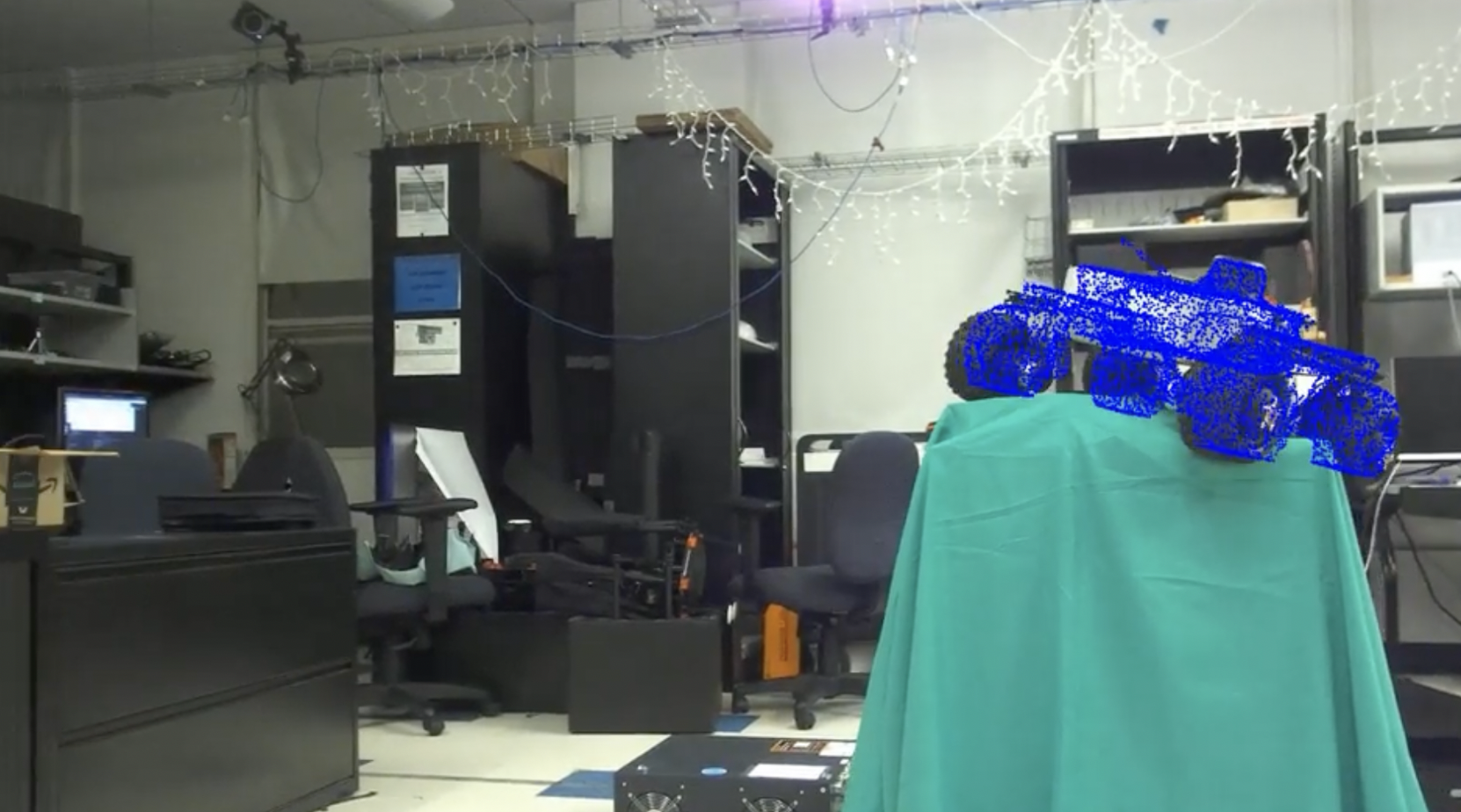}
    \caption{An example of pointcloud labeling with the leo rover model}
    \label{fig:pointcloud}
\end{figure}
We manually create high-quality 3D mesh models for each object. The models correctly represent the object's geometry and texture. We load the 3D scene point cloud and object models into labelling software and manually annotate the 6D pose for all visible objects in the first RGB-D frame. Poses are refined using ICP registration between models and point clouds. For subsequent frames, we run a pose tracking algorithm to automatically propagate the labels using point cloud registration and render the overlay. (Figure \ref{fig:pointcloud}) A human reviewer verifies the quality and manually fixes any erroneous frames. We also generate per-pixel semantic segmentation labels for each RGB-D frame through the projection of the object models using the 6D poses.

The end result is a dataset with Leo Rover captured in 6 real-world scenes with pixel-level semantic segmentation and accurate 6D poses for all objects in all frames. This pipeline allows us to efficiently generate massive amounts of annotated data. This multi-stage pipeline combining calibration, 3D alignment, automatic propagation and manual review allows us to efficiently annotate massive volumes of accurate ground truth data.

\subsection{Rover Tracking Neural Networks}
For 3D localization, robust systems can be classified as either predicting the poses of known objects in specific scenes (closed-set algorithms) \cite{densefusion,xiang_posecnn_2018,he_ffb6d_2021,he_pvn3d_2020} or generalizing the prediction to a broader scope (open-set algorithms) \cite{he_fs6d_2022}. However, open-set algorithms typically have less accuracy due to the lack of information for the target object as well as the large domain gap between the synthetic dataset and the real scenes. As a result, we opted to develop and employ a closed-set algorithm for our robot-specific interaction task.

\subsubsection{DTTDNet Model Architecture}
\begin{figure*}[ht]
    \centering
    \includegraphics[width=\linewidth]{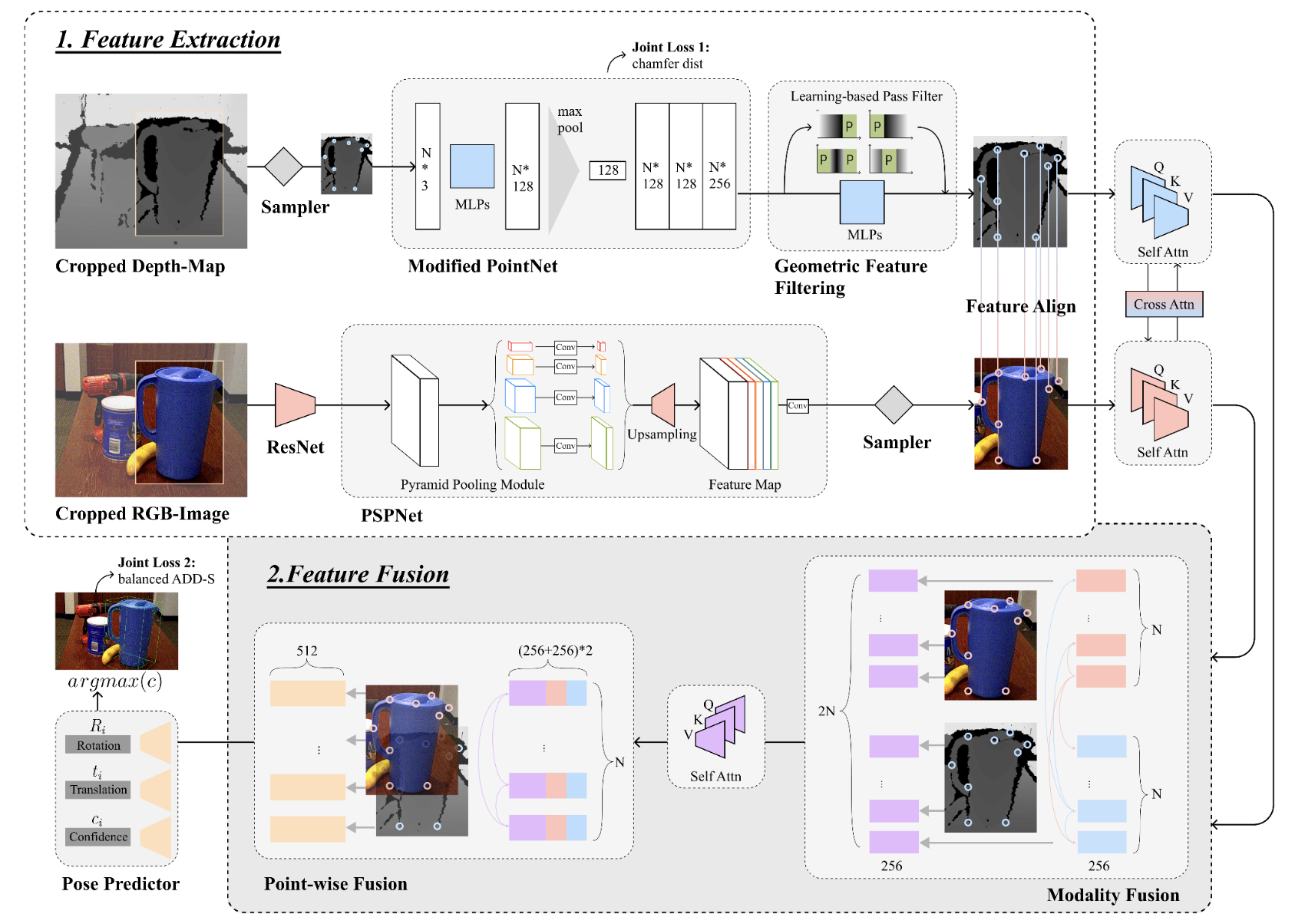}
    \caption{Model Architecture Overview. 
    The pipeline starts with performing object detection with the YOLO \cite{YOLO} network. Then the cropped RGB images, together with segmented depth maps are fed to the two feature extractors. The point cloud from the depth map and RGB colors are encoded and integrated point-wise. The embeddings then go through an attention-based two-stage fusion. Finally, point-wise prediction, rotation and translation are output from the pose predictor.}
    \label{fig:model_arch}
\end{figure*}

Here, we introduce the DTTDNet for performing 6D pose estimation of robots. Figure ~\ref{fig:model_arch} illustrates the model architecture. The approach involves two stages: detection and pose estimation.

In the detection stage, YOLOv8 \cite{YOLO}, a SOTA model initially trained on the COCO \cite{coco_2015} and ImageNet datasets \cite{deng_imagenet_2009}, is used to identify areas in each image containing the target object, which, in this context, is a robot. A 2D bounding box delineating such an area allows for cropping the image to include only the robot, which is then processed in the next stage. In our work, we fine-tuned YOLOv8 using the DTTD3 dataset \cite{Sajid_2023}. In the pose estimation stage, we adopt DTTDNet, an RGBD-based transformer network developed by Huang et al. \cite{huang_robust_2023}, renowned for its robust object pose estimation and localization capabilities. DTTDNet processes segmented depth images and cropped RGB images. It uses two feature extractors: one transforms segmented depth pixels into 3D point clouds, and the other, an image embedding network, handles RGB features. The extracted features are then encoded and integrated on a point-wise basis. These fused features serve as input of an attention-based two-stage fusion process, encompassing both modality fusion and point-wise fusion \cite{huang_robust_2023}. The final component of DTTDNet is a pose predictor, which provides point-wise predictions along with rotation and translation data. The final 6DoF pose estimation result is determined based on an unsupervised confidence scoring system \cite{huang_robust_2023}.

\subsubsection{Depth Data Robustifying with Two Loss Functions}
Specifically in the DTTDNet, two modules are used to improve the point-cloud encoder's ability to handle noisy sensor data. First, a Chamfer Distance Loss (CDL) is used to address the issue of depth data corruption due to noise and errors in data collection. It operates by comparing a decoded point set from the embedding with a reference point set (from a depth map or an object model) to guide the decoder. The second module is the Geometric Feature Filtering (GFF). GFF includes a Fast Fourier Transform (FTT) along with a single layer of MLP and an inverse-FFT. GFF handles non-Gaussian noise, and helps select significant features from noisy input signals. Together, these modules significantly improve the robustness of depth data processing in the presence of noise and low resolution, enhancing the overall performance of the DTTDNet system \cite{huang_robust_2023}.

\subsection{Experiment Process}
As a validation of our data collection process, we run two experiments by separating our dataset differently and training the pose estimation pipeline introduced in DTTD2 \cite{huang_robust_2023} independently for 2 times. 

In each experiment,  we used 17 out of 18 real-world scenes, comprising about 5,000 image frames, as training data. Additionally, we created 30,000 synthetic frames featuring various postures of Leo Rover against a dark background, showcasing both its rigid and non-rigid properties.

\subsubsection{YOLOv8 Training}
We trained the YOLOv8 network using bounding boxes derived from manually annotated point clouds. The training used both synthetic and real-world data over 500 epochs, with epoch 500 proving optimal.

\subsubsection{DTTDNet Training}
We utilize our dataset to train a simple DTTDNet as the pose estimator, which is based on pre-trained layers for feature extraction, modality fusion, and point-wise fusion. 

\begin{figure}
    \centering
    \includegraphics[width=\linewidth]{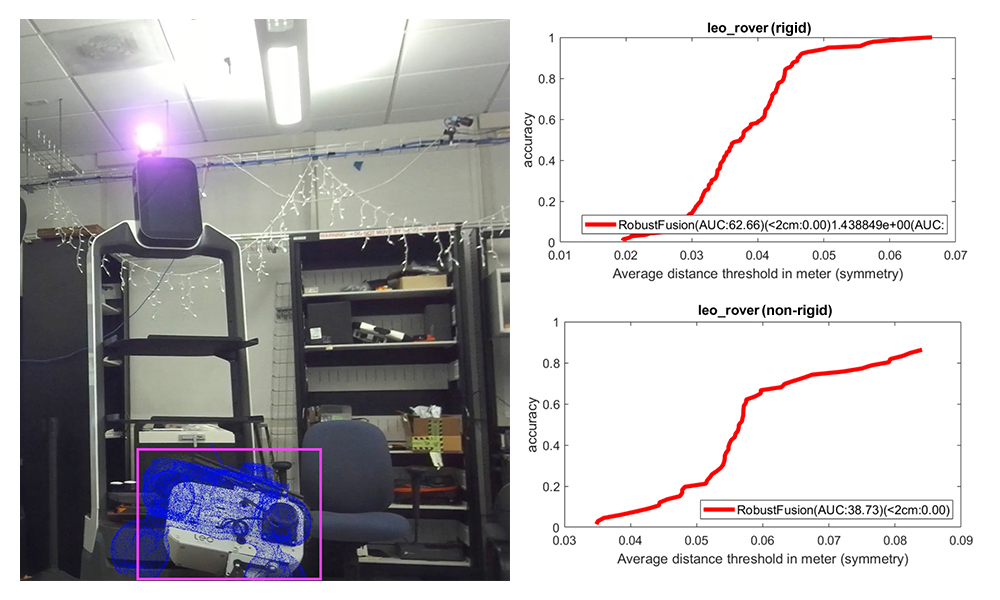}
    \caption{Output evaluation from the pose estimation pipeline. The frame represents the detected bounding box from the YOLOv8 network, and the point cloud represents the pose estimation result from the DTTDNet. ADD-S metric is used to evaluate the model's performance on the Leo Rover for both rigid-body and non-rigid-body testing scenes.}
    \label{fig:prediction}
\end{figure}

\subsubsection{Experiment Analysis}

Trained with the primitive dataset, the model becomes able to conduct rough pose estimation of the robot. As shown in figure ~\ref{fig:prediction}, the model's point-cloud predictions closely align with the Leo Rover, showing substantial robustness even with the non-rigid-body robot. We showcased this prediction pipeline at Jacobs Winter Design Showcase and performed quantitative analyses on our dataset. Employing the ADD-S metric \cite{xiang_posecnn_2018} for evaluation, the model's performance on a rigid-body testing scene with an Area-Under-Curve (AUC) of 62.66 is better than its performance on a non-rigid-body testing scene (38.73 AUC).This differential suggests a significant challenge in robust pose estimation for non-rigid objects.

\section{Challenges and Limitations}
In this study, we explored the challenges and limitations associated with developing enhanced human-robot control and interaction mechanisms. Our findings identified several areas requiring further refinement to improve the rover tracking algorithm and its applicability in real-world scenarios. Furthermore, our research highlighted that the tracking dataset itself also faces limitations when applied under actual conditions. Additionally, for the non-intrusive interaction, we discussed the hallucination nature of LLMs which may lead to inaccurate function execution. 

\subsection{Data Collection:} One of the primary challenges encountered in this study was related to data collection. The motion capture system used for ground truth data acquisition exhibited noise, which necessitated refinement to improve accuracy. Additionally, ensuring consistent data quality across different lighting conditions and robot configurations proved to be a significant challenge. Addressing these issues is crucial for obtaining reliable and representative data for training and evaluation.

\subsection{Network Training:} The limited diversity of our dataset impacted the model's ability to generalize to new scenarios. A simple network architecture was employed, which led to suboptimal performance in complex scenes. Enhancing the dataset's diversity and exploring more sophisticated network architectures are essential steps toward improving the model's robustness and performance.

\subsection{Dataset for NASA SUITS:} Our initial DTTD dataset, which includes 5000 frames of the Leo Rover, yielded unstable and jittery tracking results. We encountered two primary challenges: the dataset's size limitation and the discrepancies between the data collection environment and the rover's actual operating environment. To address these issues, we expanded the dataset by incorporating 8000 additional frames derived from real competition videos of the rover, which were used for 2D labeling.

\subsection{Hallucinations in LLMs:} Hallucinations in LLMs often involve the misinterpretation of function call parameter values, which can result in inaccurate function execution. While setting the exploration temperature to zero can help limit such errors by reducing randomness in the response generation, it does not entirely eliminate the possibility of inaccuracies. Consequently, there is a need for close supervision of the outputs and active monitoring of results, ideally with the assistance of human experts to ensure reliability and accuracy.

\section{Future directions of research}
As we look to the future, the integration of augmented reality (AR) into 3D object tracking could greatly enhance the interactivity and precision of robotic operations. Here are several key research directions:

\textbf{Enhanced AR Interfaces:}
\begin{itemize}
    \item \textit{Adaptive Interfaces: } Develop AR interfaces that adapt to user preferences, context, and environmental conditions, optimizing information display for clarity and ease of use.
    \item \textit{Contextual Data Overlay: } Provide contextual information overlays, such as operational stats or historical data for Leo Rover on the 3D model.
\end{itemize}

\textbf{Dataset Enhancement:}
\begin{itemize}
    \item \textit{Increase Diversity:} Expand the dataset to include more diverse lighting conditions, camera angles, and backgrounds. Incorporate scenes with varying degrees of clutter and occlusion to challenge the model's robustness.
    \item \textit{Real-World Scenarios:} Extend the dataset to cover more complex scenes involving dynamic interactions between robots and their environment. Include a wider variety of robot types and configurations to represent real-world applications.
\end{itemize}

\textbf{Network Improvements:}
\begin{itemize}
    \item \textit{Complex Architectures:} Explore deeper and more sophisticated network structures for enhanced pose prediction accuracy. Investigate the integration of attention mechanisms and transformer models for improved feature extraction.
    \item \textit{Movable Parts Prediction:} Develop methods to separately predict the pose of movable parts in non-rigid-body robots. Implement constraints and relationships between different robot parts for coherent pose estimation.
\end{itemize}

\textbf{Optimization and Robustness:}
\begin{itemize}
    \item \textit{Speed Optimization:} Implement techniques such as model pruning, quantization, and hardware acceleration to achieve real-time performance. Explore lightweight convolutional neural networks for faster inference.
    \item \textit{Noise Robustness:} Enhance the model's ability to handle sensor noise, especially in depth data. Investigate advanced filtering techniques and the use of generative adversarial networks (GANs) to generate synthetic training data for robust models.
\end{itemize}

\section{Conclusion}
Through this project, we presented an in-depth exploration of AR interactions, specifically focusing on robot-related tasks in space. Leveraging the power of LLMs, we developed a non-intrusive voice-controlled HMD user interface. In addition, to improve the precision of interaction with moving robots, we highlight our work with a comprehensive Digital-Twin Tracking Dataset, providing 18 annotated scenes totaling 5000 frames with varied environments and capturing scales of the Leo Rover.  The dataset was then tested using the prior
pose estimation pipeline \cite{huang_robust_2023}, exhibiting promising results. Through our work, we outline future directions that elevate the capabilities of AR in industrial and research applications, focusing on enhancing user interfaces and improving the adaptability of pose estimation technologies.

Looking ahead, the future of robotics pose estimation research as informed by our work, suggests several exciting directions. Enhancing the dataset's diversity and complexity remains a pivotal goal, aiming to further refine the generalizability of pose estimation models. We propose to augment our dataset with additional scenes that capture a broader variety of robotic forms and operational environments.



\printbibliography

\end{document}